\documentclass[sigconf]{acmart}
\usepackage{multirow}
\usepackage{balance}
\usepackage{algorithm}
\usepackage{algpseudocode}
\usepackage{amsmath}

\usepackage{algorithmicx}
\usepackage{algpseudocode}
\usepackage{float}
\usepackage{tabularray}
\usepackage{hyperref}
\hypersetup{hidelinks,
	colorlinks=true,
	allcolors=black,
	pdfstartview=Fit,
	breaklinks=true}
\usepackage{fancyhdr}
\renewcommand\footnotetextcopyrightpermission[1]{}
\AtBeginDocument{%
  \providecommand\BibTeX{{%
    \normalfont B\kern-0.5em{\scshape i\kern-0.25em b}\kern-0.8em\TeX}}}

\begin{document}

\title{Pixel Adapter: A Graph-Based Post-Processing Approach for Scene Text Image Super-Resolution}
%Robust Scene Text Image Super-Resolution Network

\author{Wenyu Zhang}
\authornote{Co-first authors with equal contribution to refining the theory and experimental design.}
\orcid{0009-0001-1457-1707}
\affiliation{%
  \institution{University of Science and Technology of China}
  \city{Hefei}
  \country{China}}
\email{wenyuz@mail.ustc.edu.cn}

\author{Xin Deng}
\authornotemark[1]
\orcid{0009-0004-6099-7516}
\affiliation{%
  \institution{University of Science and Technology of China}
  \city{Hefei}
  \country{China}}
\email{dx1998@mail.ustc.edu.cn}

\author{Baojun Jia}
\orcid{0009-0000-8841-5419}
\affiliation{%
  \institution{University of Science and Technology of China}
  \city{Hefei}
  \country{China}}
\email{june1997@mail.ustc.edu.cn}

\author{Xingtong Yu}
\orcid{0000-0002-2884-8578}
\affiliation{%
  \institution{University of Science and Technology of China}
  \city{Hefei}
  \country{China}}
\email{yxt95@mail.ustc.edu.cn}

\author{Yifan Chen}
\orcid{0009-0008-5637-2758}
\affiliation{%
  \institution{China Merchants Bank}
  \city{Chengdu}
  \country{China}}
\email{stormand@cmbchina.com}

\author{Jin Ma}
\orcid{0009-0000-4601-6372}
\affiliation{%
  \institution{China Merchants Bank}
  \city{Chengdu}
  \country{China}}
\email{majin@cmbchina.com}

\author{Qing Ding}
\authornote{Corresponding authors.}
\orcid{0000-0002-3752-1163}
\affiliation{%
  \institution{University of Science and Technology of China}
  \city{Hefei}
  \country{China}}
\email{dingqing@ustc.edu.cn}

\author{Xinming Zhang}
\authornotemark[2]
% \authornote{Corresponding authors.}
\orcid{0000-0002-8136-6834}
\affiliation{%
  \institution{University of Science and Technology of China}
  \city{Hefei}
  \country{China}}
\email{xinming@ustc.edu.cn}

\renewcommand{\shortauthors}{Wenyu Zhang et al.}

\begin{abstract}
Current Scene text image super-resolution approaches primarily focus on extracting robust features, acquiring text information, and complex training strategies to generate super-resolution images. However, the upsampling module, which is crucial in the process of converting low-resolution images to high-resolution ones, has received little attention in existing works. To address this issue, we propose the Pixel Adapter Module (PAM) based on graph attention to address pixel distortion caused by upsampling. The PAM effectively captures local structural information by allowing each pixel to interact with its neighbors and update features. Unlike previous graph attention mechanisms, our approach achieves 2-3 orders of magnitude improvement in efficiency and memory utilization by eliminating the dependency on sparse adjacency matrices and introducing a sliding window approach for efficient parallel computation. Additionally, we introduce the MLP-based Sequential Residual Block (MSRB) for robust feature extraction from text images, and a Local Contour Awareness loss ($\mathcal{L}_{lca}$) to enhance the model's perception of details. Comprehensive experiments on TextZoom demonstrate that our proposed method generates high-quality super-resolution images, surpassing existing methods in recognition accuracy. For single-stage and multi-stage strategies, we achieved improvements of 0.7\% and 2.6\%, respectively, increasing the performance from 52.6\% and 53.7\% to 53.3\% and 56.3\%. The code is available at https://github.com/wenyu1009/RTSRN.
\end{abstract}







\maketitle

\section{Introduction}

Scene text image super-resolution (STISR) \cite{6]DBLP:conf/eccv/WangX0WLSB20} is essential for scene text recognition, as it generates high-resolution images for subsequent text analysis. This technique finds applications in a variety of domains, including card recognition \cite{3]DBLP:journals/eswa/KhareSCLLHB19}, sign recognition \cite{1]DBLP:journals/cviu/FangFYCC04}, autonomous driving \cite{2]DBLP:journals/tits/ZhangDPFW21}, and others \cite{4]20190906545388,5]DBLP:journals/pr/SanchezRTVV19}. Recent methods in text image super-resolution \cite{12]DBLP:conf/cvpr/MaLZ22} have employed various techniques to enhance model performance, such as high-performance feature extraction \cite{6]DBLP:conf/eccv/WangX0WLSB20}, obtaining multimodal information through auxiliary models, and complex training strategies \cite{12]DBLP:conf/cvpr/MaLZ22, 15DBLP:journals/corr/abs-2204-14044}. Previous methods, such as STT \cite{13DBLP:conf/cvpr/ChenLX21} proposed Transformer-Based Super-Resolution Network (TBSRN) to capture sequential information, while TATT \cite{12]DBLP:conf/cvpr/MaLZ22} and C3-STISR \cite{15DBLP:journals/corr/abs-2204-14044} use auxiliary models to recognize text information and guide text image generation. TPGSR \cite{11]DBLP:journals/corr/abs-2106-15368} and C3-STISR \cite{15DBLP:journals/corr/abs-2204-14044} employ complex training strategies to further improve the performance of their models. 

\begin{figure}[t]
% \vspace{5mm}
  \centering
   %\includegraphics[width=0.8\linewidth]{egfigure.eps}
   \includegraphics[width=\linewidth]{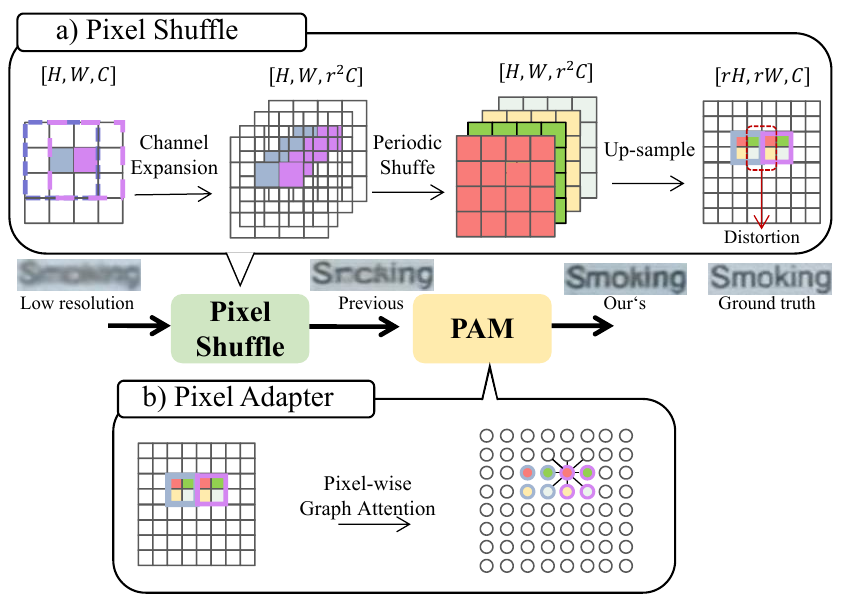}
   \caption{a) Pixel shuffle, a commonly used upsampling module in the field of image super-resolution. b) Pixel adapter, a graph-based pixel interaction module.}
   \label{fig:1}
\end{figure}

It can be found that these previous works mainly focused on various modules for enhancing super-resolution performance but overlooked the upsampling module and its influence on image quality. Traditional methods like nearest neighbor interpolation, bilinear interpolation, and bicubic interpolation result in problems such as blurring, jagged edges, and loss of high-frequency details. More advanced techniques, including deep learning-based methods like Super-Resolution Convolutional Neural Networks (SRCNN) \cite{55DBLP:conf/eccv/DongLHT14}, Pixel Shuffle \cite{56DBLP:conf/cvpr/ShiCHTABRW16}, and Deep Fourier Up-Sampling \cite{57DBLP:journals/corr/abs-2210-05171}, attempt to address these issues and demonstrate significant improvements in preserving image details. However, challenges still remain. As shown in Fig.\ref{fig:1}, the widely adopted Pixel Shuffle method aggregates the low-resolution image neighboring pixels by using a convolutional kernel of size 3 and stride 1, ensuring ordered correlations within the 4-pixel block obtained after the up-sample. However, the pixel relationships between groups are ignored, resulting in meaningless noise in adjacent pixels between groups. When these 4-pixel units are stacked to obtain the super-resolution image, leading to the degradation of medium-to-high frequency information and distortion between pixels. It exhibits jagged edges and image blur. \textbf{Interestingly, although upsampling aims to achieve super-resolution images, it may inadvertently introduce certain issues that degrade image quality.}

In this paper, we propose an efficient pixel-wise graph attention module, called the Pixel Adapter Module (PAM), to address the issue of pixel distortion caused by upsampling in image super-resolution. This module utilizes a graph attention mechanism \cite{58DBLP:conf/iclr/VelickovicCCRLB18} to allow each pixel to interact with its neighbors, capture local structural information, and update features effectively to resolve pixel distortion. In comparison, the global attention mechanism \cite{59DBLP:conf/nips/VaswaniSPUJGKP17} struggles with handling local inconsistencies and incurs high computational and memory costs. On the other hand, the block-based attention mechanism \cite{20DBLP:conf/cvpr/VaswaniRSPHS21} lacks the required granularity.

Existing pixel-wise graph attention methods \cite{22DBLP:conf/mm/ZhangDHML21}, which treat pixels as nodes in the graph and interact with surrounding pixels, have shown potential in addressing local pixel distortion problems. Nonetheless, these methods still rely on an adjacency matrix to guide the update of node features, leading to sparsity in the matrix, resulting in significant memory consumption and slower processing speed. To overcome this limitation, our proposed method eliminates the dependency on sparse adjacency matrices and introduces an innovative sliding window approach to efficiently capture neighborhood relationships in the image in a parallel manner. Specifically, neighbors in the image are defined as the surrounding pixels within a 3x3 matrix, with the central point acting as the target node and the surrounding points as neighbors (8 neighbors). Consequently, the sliding window enables interactions between nodes and their neighbors, promoting parallel computation. This approach significantly reduces memory consumption and improves processing speed. Compared to previous methods \cite{22DBLP:conf/mm/ZhangDHML21}, this technique achieves 2-3 orders of magnitude improvement in efficiency and memory utilization. This advancement makes it possible to utilize pixel-wise graph attention on large-scale images, providing an efficient solution to the pixel distortion problem caused by upsampling.

In addition, robust feature extraction from text images serves as the foundation for subsequent processing. Due to the sequential nature of the text, characters exhibit strong correlations in spatial arrangement and context. To capture these interdependencies, we propose the MLP-based Sequential Residual Block (MSRB), which employs both horizontal and vertical MLP operators to extract image features, complemented by LSTM modeling for sequence information. Furthermore, to enhance the model's perception of details and high-frequency information, we propose a Local Contour Awareness loss ($\mathcal{L}_{lca}$), utilizing Histogram of Oriented Gradient (HOG) \cite{60DBLP:conf/cvpr/DalalT05} features as reconstruction targets to maintain invariance to geometric and optical distortions. Combining these improvements, we introduce a robust model called the Robust Text Super-Resolution Network (RTSRN), with experiments demonstrating the effectiveness of the proposed modules and the model's generalizability.

Our contributions are summarized below:
\begin{itemize}   
\item We propose a pixel adapter module based on pixel-wise graph attention to mitigate the issue of information loss and discontinuity in super-resolution feature maps, achieving 2-3 orders of magnitude improvement in efficiency and memory utilization compared to previous pixel-wise graph attention mechanisms.
\item We propose an MSRB for STISR tasks with stronger feature extraction ability and better knowledge learning from auxiliary models.
\item We propose a local contour awareness loss, which guides the model to generate super-resolution images with more details and high quality.
\end{itemize}

\section{Related work}
In this section, we discuss recent advancements in vision backbone, STISR methods, and common loss functions.

\subsection{Vision Backbone}

Recently, the advancement of the vision backbone has been driven by various innovative approaches. ResNet \cite{28DBLP:conf/codit/FerjaouiCAZ22}, which introduced shortcut connections in 2015, remains widely adopted. However, newer methods like Vision Transformer \cite{20DBLP:conf/cvpr/VaswaniRSPHS21,21DBLP:journals/corr/abs-2107-00782,22DBLP:conf/mm/ZhangDHML21,25DBLP:conf/nips/ParmarRVBLS19,27DBLP:conf/iclr/DosovitskiyB0WZ21,1920210401013} and Vision MLP \cite{18DBLP:journals/corr/abs-2105-01601,1920210401013} have demonstrated strong performance in visual tasks. To overcome the limitations of these backbones, such as their reliance on extensive training with large datasets, novel techniques have emerged. PGANet \cite{22DBLP:conf/mm/ZhangDHML21} combines ResNet with pixel-wise attention for improved efficiency and accuracy, while HaloNet \cite{20DBLP:conf/cvpr/VaswaniRSPHS21} proposes a non-centered blocked local attention mechanism. Additionally, WaveMLP \cite{20DBLP:conf/cvpr/VaswaniRSPHS21} employs a dynamic aggregation block based on wave-represented image features, and ConvNeXt \cite{22DBLP:conf/mm/ZhangDHML21} modernizes ResNet to effectively compete with Transformers.

\subsection{Scene Text Image Super-Resolution}
In the early stages of scene text image processing, some techniques employed a general super-resolution model to directly enhance the resolution of images \cite{6]DBLP:conf/eccv/WangX0WLSB20,9]DBLP:conf/rivf/TranH19,10]DBLP:journals/pami/ShiBY17}. SRResNet \cite{51DBLP:conf/cvpr/LedigTHCCAATTWS17} introduced the use of generative adversarial networks to generate more distinct images. In \cite{9]DBLP:conf/rivf/TranH19}, a Laplacian-pyramid backbone was utilized to upsample low-resolution images. TSRN \cite{6]DBLP:conf/eccv/WangX0WLSB20} was proposed, incorporating a horizontal and vertical BLSTM to capture sequential information in text images. More recent methods use auxiliary models to extract text information from images to guide the super-resolution process. For instance, TATT \cite{12]DBLP:conf/cvpr/MaLZ22} leverages a recognition model to acquire prior knowledge of the text, guiding the model to restore the shape and texture of the text in the image. C3-STISR \cite{15DBLP:journals/corr/abs-2204-14044} utilizes the feedback from a recognizer, combined with visual and linguistic information, to generate more recognizable images. Additionally, multi-stage training strategies have been proposed in C3-STISR and TPGSR \cite{52JianqiMa2021TextPG} to achieve improved performance.

\subsection{Loss Function In STISR}

In STISR, Wang et al. \cite{6]DBLP:conf/eccv/WangX0WLSB20} use the gradient fields of high-resolution images and super-resolution images to guide the model training process. The sharpening of character boundaries is employed to make the characters more distinct and legible. TextSR \cite{30DBLP:journals/corr/abs-1801-04381} employs a text perceptual loss to train the super-resolution network, thereby giving priority to the text content rather than the background. Text Gestalt \cite{14DBLP:journals/corr/abs-2112-08171} introduces the Stroke-Focused Module (SFM) to concentrate on the stroke-level internal structures of characters in text images. STT \cite{13DBLP:conf/cvpr/ChenLX21} employs a Position-Aware Module and a Content-Aware Module to emphasize the position and content of each character, leveraging a pre-trained Transformer on a synthesized text dataset \cite{53DBLP:conf/cvpr/GuptaVZ16, 54DBLP:journals/ijcv/JaderbergSVZ16}. TATT \cite{12]DBLP:conf/cvpr/MaLZ22} proposes a text structure consistency loss to refine the visual appearance by enforcing structural consistency on the reconstructions of regular and deformed texts.

\begin{figure*}[!ht]
\vspace{-1 em}
  \centering
   %\includegraphics[width=0.8\linewidth]{egfigure.eps}
   \includegraphics[width=0.9\linewidth]{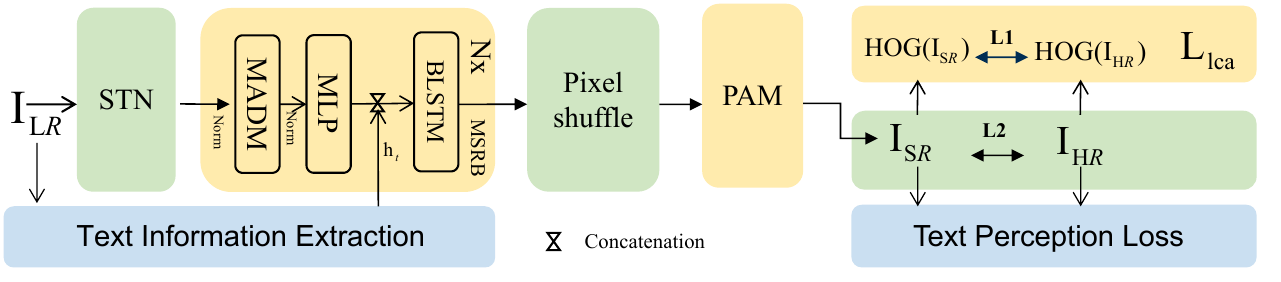}
   \caption{Architecture of RTSRN, the yellow areas represent the MSRB, PAM, and $\mathcal{L}_{lca}$ that we proposed.}
   \label{fig:architecture}
\end{figure*}

\section{Methodology}

In this section, we first provide an overview of the model architecture and the pipeline of STISR in Sec.\ref{3.1}. Then, we sequentially introduce the methods proposed in this paper, following the data processing flow. In Sec.\ref{3.2}, we introduce the feature extraction module MLP-based Sequential Residual Block. Next, in Sec.\ref{3.3}, we present the Pixel Adapter Module. Finally, in Sec.\ref{3.4}, we discuss the Local Contour Awareness Loss utilized in our model training.

\subsection{Overall Architecture}
\label{3.1}
The architecture of our proposed model is illustrated in Fig. \ref{fig:architecture}. The yellow components are MSRB, PAM, and $\mathcal{L}_{lca}$. We selected C3-STISR as our baseline \footnote{We have established the baseline with reference to C3-STISR. It's worth noting that the official code of the paper has removed the visual clues.}. By replacing MSRB with RSB and removing PAM and $\mathcal{L}_{lca}$, we arrived at the baseline model architecture.

The pipeline for text image super-resolution processing is outlined below. The input to the model is the low-resolution image $I_{LR} \in \mathbb{R}^{H \times W \times 3}$, where $H$, $W$ are the height and width of the image, and $3$ is the number of channels. The model processes the input through two paths: the super-resolution path and the text information extraction path. In the super-resolution path, the spatial deformation of the image is processed using STN \cite{7]DBLP:conf/nips/JaderbergSZK15} to obtain the feature map $i_{lr}$. The feature then enters the MSRB $f_{msrb}$. 

Simultaneously, the low-resolution image $I_{LR}$ is input into the text information extraction path to extract text clues. The text probability $h_{rec}$ of the text in the image is obtained using a CRNN \cite{10]DBLP:journals/pami/ShiBY17}. However, recognition models have limitations in terms of modal compatibility and recognition accuracy. Therefore, following C3-STISR \cite{15DBLP:journals/corr/abs-2204-14044}, we used a pre-trained character-level language model to correct the predicted text and obtain linguistic cues $h_{ling}$. These text clues are then fused in the fusion module to form the final text clues  $h_t$, which are transformed into a unified pixel feature map of $C \times H \times W$. \textbf{Please refer to Appendix.\ref{appdenx.1} for more details.}

Next, the text clues are combined with the image features obtained from the feature extraction block to form multimodal data, which are then encoded using a bidirectional LSTM (BLSTM) to capture contextual information. Utilizing fused features for pixel shuffle generates super-resolution feature maps, and to mitigate image discontinuities caused by the upsampling process, the PAM is employed for further optimization and smoothing of pixels, resulting in higher-quality super-resolution images. The above process can be formulated as $I_{SR}= f_{pam}(f_{ps}(f_{msrb}([STN(I_{LR}),h_{t}])))$, where [·] represents concatenation along the channel dimension. To compensate for the absence of high-frequency information in low-resolution images, our model training incorporates the $\mathcal{L}_{lca}$ Loss, which enhances the details and overall visual quality. \textbf{Owing to the limited space of the paper, and other loss functions refer to the Appendix.\ref{appdenx.2}.}

% Owing to the limited space of the paper, we will introduce MSRB, PAM, and $\mathcal{L}_{lca}$ in the order of data flow within the model. For details on the extraction of the text clues $h_t$ and other loss functions used during training, please refer to the supplementary materials.

\subsection{MLP-based Sequential Residual Block}
\label{3.2}
% The features obtained by SRB will significantly impact the generation of subsequent super-resolution images, and the simple two-layer CNN is inadequate in terms of feature representation. Recently, vision backbones based on CNN, MLP, and Attention have made significant progress. In order to enhance both performance and efficiency, we decided to use the basic block of ConvNeXt \cite{17DBLP:conf/cvpr/0003MWFDX22}, WaveMLP \cite{1920210401013}, and HaloNet \cite{20DBLP:conf/cvpr/VaswaniRSPHS21} to replace the two-layer CNN in the SRB for image feature extraction. These works, apart from different operators, adopt a basic block design similar to that of Transformers, which also proves that key components contribute to performance improvements, such as fewer normalization layers, fewer activation functions, and larger convolutional kernels. Based on this, we propose an MLP-based Sequential Residual Block. 

% This choice has two main reasons: 1) MSRB effectively extracts features from the horizontal, vertical, and channel directions and uses a dynamic aggregation module to merge information from multiple directions, enabling robust processing of text images from various directions. 2) During model training, multiple auxiliary models are used to detect text clues and character regions in scene text images. MLP with lower inductive bias is able to learn from these auxiliary models more effectively.

The features obtained by the feature extraction module significantly impact the subsequent super-resolution image generation. The SRB, which utilizes a simple two-layer CNN for feature extraction, falls short in terms of feature representation. Recently, there have been remarkable advancements in visual backbone networks based on CNN \cite{17DBLP:conf/cvpr/0003MWFDX22}, MLP \cite{1920210401013}, and Attention \cite{20DBLP:conf/cvpr/VaswaniRSPHS21}, both in terms of operators and architectures. Taking into account the characteristics of text images, where text information primarily exists in the horizontal and vertical directions, we propose the MSRB, drawing inspiration from existing advanced feature extraction modules.

\begin{figure}[t]
  \centering
  \vspace{-1 em}
   %\includegraphics[width=0.8\linewidth]{egfigure.eps}
   \includegraphics[width=0.7\linewidth]{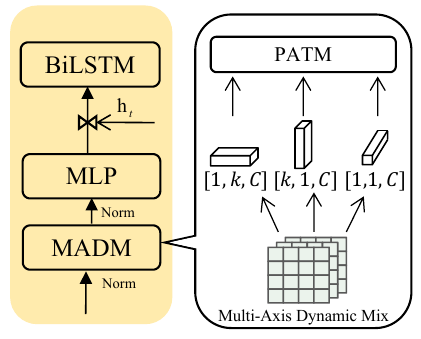}
   \caption{MLP-based Sequential Residual Block. It utilizes MLP to extract robust features from three directions and uses dynamic fusion modules for fusion.}
   \label{fig:3}
\end{figure}

The specific structure of MSRB is illustrated in Fig.\ref{fig:3}. The input features are fed into the Multi-Axis Dynamic Mix layer, which utilizes MLPs in the horizontal, vertical, and channel directions to extract features. The resulting three feature maps are then inputted into the Dynamic Fusion module to obtain image features. Next, the extracted features are input into an MLP (Multi-Layer Perceptron) feedforward network. It's important to note that both MADM and MLP have residual structures. These features are concatenated with the text clues extracted from Text Information Extraction (see Appendix. \ref{appdenx.1}) and inputted into a Bidirectional LSTM, as follows:

\begin{equation}
\boldsymbol{MSRB}(\boldsymbol{i}_{lr},\boldsymbol{h}_{t})=\boldsymbol{BLSTM}([\boldsymbol{MLP}(\boldsymbol{MADM}(\boldsymbol{i}_{lr})),\boldsymbol{h}_{t}]),
\end{equation}

We adopt PATM \cite{1920210401013} as a dynamic fusion module, it treating image features as waves with amplitude and phase, $\tilde{\boldsymbol{z}}=\left|\boldsymbol{z}\right| \odot e^{i \boldsymbol{\theta}}$, where, $\lvert \boldsymbol{z}_j \rvert$ and $e^{i \boldsymbol{\theta}_j}$ represent the amplitude and phase information of each component, respectively. $\odot$ is element-wise multiplication. We made modifications to the super-resolution task by aggregating features of the whole image, rather than dividing the image into patches. For simplicity, the absolute value calculation for the amplitude and phase is directly produced by the Channel-FC component. Therefore, the PATM is as follows:
\begin{equation}
\boldsymbol{PATM}(\boldsymbol{i}_{lr})=\boldsymbol{MLP}(\sum_k W_{k}^t z_k \odot \cos \theta_k+W_{k}^i z_k \odot \sin \theta_k ),
\end{equation}
where $W$ represents the learnable weights, and $K$ refers to dynamic aggregation performed in several directions. The variable $z_k$ is obtained from $\boldsymbol{z}=\text { Channel-FC }\left(\boldsymbol{i}_{lr}, W^c\right)$. The phase information $\theta_k$ is generated based on the input image feature $\boldsymbol{i}_{lr}$ through $\boldsymbol{\theta}_j=\Theta\left(x_j, W^\theta\right)$, where $W^\theta$ is a learnable parameter. Finally, the output is processed by an $\boldsymbol{MLP}$ to enhance its representation ability.

\subsection{Pixel Adapter Module}
\label{3.3}

% The main goal of upsampling in super-resolution tasks is to change the original feature map $i_{lr} \in \mathbb{R}^{H \times W \times C}$ into the corresponding  $i_{sr} \in \mathbb{R}^{H*r \times W*r \times C}$  according to the upsampling factor $r$. Pixel Shuffle is usually used as an upsampling module to obtain high-resolution feature maps through convolution and periodic shuffling. Specifically, for low-resolution images, convolution is first used to change the low-resolution feature map $i_{lr} \in \mathbb{R}^{H \times W \times C}$ to $i'_{lr} \in \mathbb{R}^{H \times W \times C*r^2}$. followed by periodic shuffling operator to change $i'_{lr} \in \mathbb{R}^{H \times W \times C*r^2}$ to $i_{sr} \in \mathbb{R}^{H*r \times W*r \times C}$. Through these two operations, the super-resolution feature map $i_{sr}$ is obtained, but it fails to consider the correlation between pixels, which results in loss of high-frequency information and blurring of details in the image.

% \begin{figure}[h!]
%   \centering
%   % \fbox{\rule{0pt}{2in} \rule{0.9\linewidth}{0pt}}
%    %\includegraphics[width=0.8\linewidth]{egfigure.eps}
%    \includegraphics[width=0.8\linewidth]{pic/fig3.pdf}
%    \caption{Different types of attention mechanisms}
%    \label{fig:3}
% \end{figure}

In this study, we employ the attention mechanism to adaptively enhance the super-resolution feature map. Existing attention mechanisms can be categorized into various forms, including global-wise, block-wise, and pixel-wise attention. In super-resolution tasks, the attention mechanism improves the upsampling process by focusing on crucial features, such as edges and textures, preserving the image structure and high-frequency information. This results in enhanced performance and reduced blurring, enabling models to generate detailed, high-quality images. The task of text image super-resolution requires low computational complexity, minimal memory consumption, and the capability to concentrate on local details for capturing high-frequency information. To fulfill these requirements, we propose PAM based on the pixel-wise graph attention mechanism, which leads to a finer-grained super-resolution feature map.

\begin{figure}[h]
\vspace{-1 em}
  \centering
   %\includegraphics[width=0.8\linewidth]{egfigure.eps}
   \includegraphics[width=0.8\linewidth]{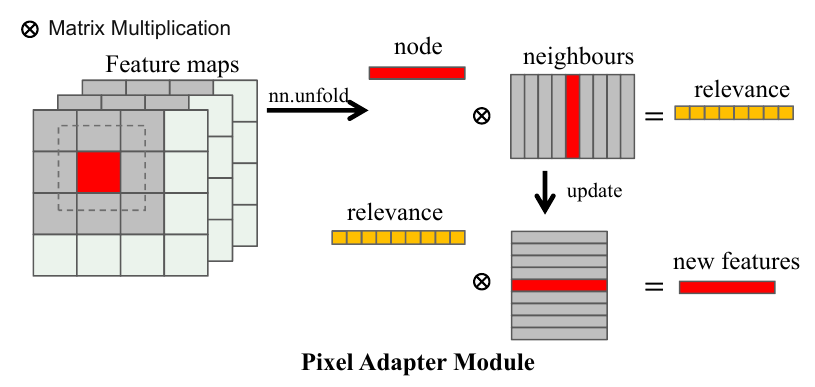}
   \caption{Efficient Pixel-wise Graph Attention Mechanism. It utilizes prior knowledge of the image, with each pixel and its surrounding neighbors considered in a window, and achieves efficient propagation of node features on the graph through the use of sliding windows.}
   \label{fig:4}
\end{figure}

This approach regards pixels as nodes within the graph, and updates their features utilizing local graph attention mechanisms \cite{23DBLP:journals/corr/abs-1710-10903}. By executing pixel smoothing grounded on the interrelations among local pixels, we attain a more refined and elaborate feature map. As shown in Fig.\ref{fig:4}, each pixel's channel values in the image are treated as the node's features. In the feature finer-grained process, each node calculates relevance with its neighboring nodes through attention and then uses the weighted sum to obtain updated features. To balance the trade-off between speed and memory usage, we adopt a convolution-like form to incorporate prior knowledge and extract local structure information between adjacent regions through overlapping windows.

Because of the regular arrangement of image pixels, the location information of the pixels is already included when we use a convolution-like window to obtain the local region. As a result, we directly calculate the relationship between the centroid and the neighboring pixels. Let $N_i$ be the set of $k$ neighbors of pixel $i$, $C$ be the number of channels in the input feature maps, $f_i^l$ and $f_j^l$ be the input features of pixels $i$ and $j$ in layer $l$, respectively. The relationship matrix between pixel $i$ and pixel $j$ in $N_i$ can be represented as follows:
\begin{equation}
    R^l=Softmax(\sum_{c=1}^{C}{\theta(f_i^l)\cdot\phi(f_j^l)/\sqrt{C})|i\in n, j\in N_i},
\end{equation}
where $Softmax$ is the softmax function for the dimension of $j$, $\theta, \phi$ is the transform MLPs, $n$ denotes the number of pixels in the input feature maps. The shape of the relationship matrix $R$ is $n \cdot k$.

Based on the relationship between the centroids and their neighboring points, we assign weights to each neighbor's features, indicating the extent of their influence on the centroids, allowing for the update of the centroid features for this local region as follows:
\begin{equation}
    f^{l+1}=\sum_{j\in N_i}{R_j^l\cdot \omega (f_j^{l})},
\end{equation}
where $\omega$ is the transform layer, $R_j^l$ is the relationship between pixel $j$ and it's neighbour $i$ in Relationship matrix $R^l$. 

Existing pixel-wise attention methods, such as SASA \cite{25DBLP:conf/nips/ParmarRVBLS19}, which employs attention masks to simulate pixel-centric neighborhoods, and PGANet \cite{22DBLP:conf/mm/ZhangDHML21}, which treat the image as a graph using an adjacency matrix, both suffer from drawbacks like high memory consumption and sluggish processing speeds. Our approach, inspired by HaloNet \cite{20DBLP:conf/cvpr/VaswaniRSPHS21}, incorporates sliding windows for parallel and efficient pixel-wise graph attention, as illustrated in Algorithm \ref{alg.1}.

\begin{algorithm}[htp]

    \caption{Pixel Adapter Algorithm}
    \label{alg.1}
    \begin{algorithmic}[1] %每行显示行号
            \Require \Call{CNN}{$\mathbf{f}$} is $1\times1$ CNN for feature transformation. \Call{Unfold}{$\mathbf{f}, \mathbf{k}, \mathbf{p=k/2}$} sliding extraction of local region blocks,$\mathbf{k}$ denotes the size of the sliding window, while $\mathbf{p}$ represents the number of extra padding elements added around the edges of the input tensor.
        \Ensure $\mathbf{f'}$ is a feature map updated based on the local graph attention mechanism.
        \Function {PAM}{$\mathtt{f}$}
            \State $\mathbf{b},\mathbf{c},\mathbf{h},\mathbf{w} \gets \mathbf{f}.shape$;
            \State $\mathbf{n} \gets \mathbf{h}*\mathbf{w}$;
            \State $\mathbf{q,k,v} \gets \Call{CNN}{\mathbf{f}}, \Call{CNN}{\mathbf{f}}, \Call{CNN}{\mathbf{f}}$;
            \State $\mathbf{q} \gets \mathbf{q}.\Call{Reshape}{\mathbf{b},\mathbf{c},\mathbf{n},\mathbf{-1}}.\Call{Permute}{0,2,3,1}$;
            \Statex \Comment{[b,c,h,w]->[b,c,n,1]->[b,n,1,c]}
           
            \State $\mathbf{k} \gets  \Call{Unfold}{\mathbf k}.\Call{Reshape}{\mathbf{b},\mathbf{c},\mathbf{-1},\mathbf{n}}.\Call{Permute}{0,3,1,2}$;
            \Statex \Comment{[b,c*k*k,n]->[b,c,k*k,n]->[b,n,c,k*k]}

            \State $\mathbf{v} \gets  \Call{Unfold}{\mathbf v}.\Call{Reshape}{\mathbf{b},\mathbf{c},\mathbf{-1},\mathbf{n}}.\Call{Permute}{0,3,2,1}$;
            \Statex \Comment{[b,c*k*k,n]->[b,c,k*k,n]->[b,n,k*k,c]}

            \State $\mathbf{att} \gets \Call{Softmax}{\mathbf{q}@\mathbf{k}/\sqrt{c}}$; 
            \Statex \Comment{[b,n,1,k*k]}
            \State $\mathbf{f'} \gets \mathbf{att}@\mathbf{v}$; 
            \Statex \Comment{[b,n,1,c]}

            \State $\mathbf{f'} \gets \mathbf f'.\Call{Squeeze}{2}.\Call{Permute}{0,2,1}.\Call{Reshape}{\mathbf{b},\mathbf{c},\mathbf{h},\mathbf{w}}$; 

            \State \Return $\mathbf f'$
        \EndFunction
        
    \end{algorithmic}
\end{algorithm}

From the perspective of computational complexity, global-wise, block-wise, and pixel-wise attention are $\mathcal{O}(n^{2}c)$, $\mathcal{O}((\frac{n}{w})^2c)$ and $O(nw^2c)$, respectively. where $n$ represents the number of points participating in the calculation, w represents the window size, and c represents the channel dimension. The computational complexity of PAM is closely tied to the sliding window. When the window size is small, its computational complexity is comparable to, or even less than, that of block-wise attention.

\subsection{Loss Function}
\label{3.4}
% \vspace{-1 em}

\begin{figure}[h]
  \centering
   %\includegraphics[width=0.8\linewidth]{egfigure.eps}
   \includegraphics[width=0.8\linewidth]{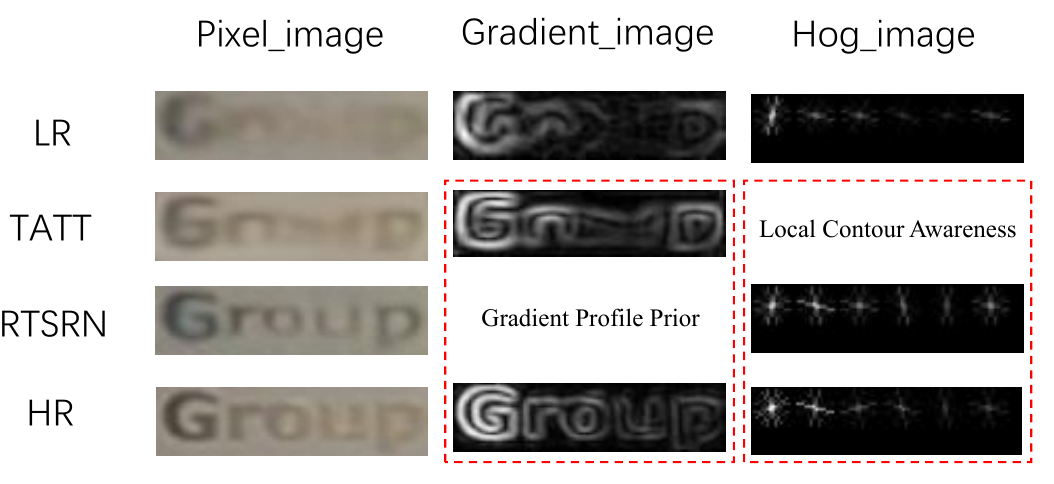}
   \caption{Reconstruction target visualization based on HOG ensures robustness against geometric and optical changes (on the right), while the gradient-based approach does not.}
   \label{fig:5}
\end{figure}

Blurred text images are often caused by factors such as insufficient lighting, camera shake, and focus issues. Previous methods have relied on the original image or image gradient profile \cite{6]DBLP:conf/eccv/WangX0WLSB20} as reconstruction targets, but these methods struggle to resist deformation and blurring caused by these factors. As shown in Fig.\ref{fig:5}, we visualize different reconstruction targets, and the gradient-based reconstruction target can resist blurring to some extent but is difficult to solve the deformation problem. The reconstructed super-resolution images still exhibit deformation and blurring. To overcome these limitations, we propose a local contour awareness loss based on image priors and utilize the Histogram of Oriented Gradients (HOG) \cite{60DBLP:conf/cvpr/DalalT05} for the super-resolution task of images. HOG is advantageous because it can effectively capture local shapes by extracting edge and gradient features, and has good invariance to geometric and optical changes due to local extraction of the image through gamma correction and gradient direction quantization. The proposed loss is as follows:
\begin{equation}
    \mathcal{L}_{\mathrm{lca}}=\left\|\mathbf{HOG}(\mathbf{I}_{\mathrm{HR}})-\mathbf{HOG}(\mathbf{I}_{\mathrm{SR}})\right\|_1,
\end{equation}

Specifically, given a pair of high-resolution (HR) and super-resolution (SR) images $(\mathbf{I}_{\mathrm{HR}},\mathbf{I}_{\mathrm{SR}})$, we aim to minimize the difference in their HOG features. The HOG operator operates on the local square units of the image, ensuring robustness against geometric and optical changes.

% The computation of HOG involves the following steps: Firstly, gradient information is extracted from the image through the application of gradient operators. These gradient values provide information about the intensity changes in the image and direction information about the edges in the image. Secondly, the gradient information is quantized into a set of predetermined orientations, typically defined in terms of a histogram of directions. Thirdly, the quantized gradient information is divided into a set of non-overlapping cells, and a histogram is computed for each cell. Finally, these local histograms are concatenated to form a feature descriptor that characterizes the structural information in the image. The HOG operator operates on the local square units of the image, ensuring robustness against geometric and optical changes. During the model training phase, the loss function is designed with reference to C3-STISR \cite{15DBLP:journals/corr/abs-2204-14044} and STT \cite{13DBLP:conf/cvpr/ChenLX21} loss, a detailed explanation of these loss functions can be found in the supplementary materials.

\begin{table*}[ht]
\vspace{-1  em}

\centering
\caption{Text recognition results of super-resolution images obtained from different models, where '-3' represents three-stage training in TPGSR, and 'mp represents multi-stage pre-training.}
\label{tab:comparison}
\begin{tabular}{c|cccc|cccc|cccc} 
\toprule
\multirow{2}{*}{Method} & \multicolumn{4}{c}{CRNN}                                     & \multicolumn{4}{c}{MORAN}                                    & \multicolumn{4}{c}{ASTER}                               \\ 
\cline{2-13}
                        & Easy          & Medium        & Hard          & Average       & Easy          & Medium        & Hard          & Average       & Easy          & Medium & Hard          & Average        \\ 
\hline
LR                      & 37.5          & 21.4          & 21.1          & 26.7          & 56.2          & 35.9          & 28.2          & 40.1          & 62.4          & 41.3   & 31.6          & 45.1           \\
HR                      & 76.4          & 75.1          & 64.6          & 72.4          & 91.2          & 85.3          & 74.2          & 84.1          & 94.2          & 87.7   & 76.2          & 86.6           \\ 
\hline
SRResNet \cite{51DBLP:conf/cvpr/LedigTHCCAATTWS17}                & 45.2          & 32.6          & 25.5          & 35.1          & 66.0          & 47.1          & 33.4          & 49.9          & 69.4          & 50.5   & 35.7          & 53.0           \\
TSRN \cite{6]DBLP:conf/eccv/WangX0WLSB20}                   & 52.5          & 38.2          & 31.4          & 41.4          & 70.1          & 55.3          & 37.9          & 55.4          & 75.1          & 56.3   & 40.1          & 58.3           \\
STT \cite{13DBLP:conf/cvpr/ChenLX21}                     & 59.6          & 47.1          & 35.3          & 48.1          & 74.1          & 57.0          & 40.8          & 58.4          & 75.7          & 59.9   & 41.6          & 60.1           \\
TPGSR \cite{52JianqiMa2021TextPG}                  & 61.0          & 49.9          & 36.7          & 49.8          & 72.2          & 57.8          & 41.3          & 57.8          & 77.0          & 60.9   & 42.4          & 60.9           \\
TATT \cite{12]DBLP:conf/cvpr/MaLZ22}                     & 62.6          & 53.4          & 39.8          & 52.6          & 72.5          & 60.2          & 43.1          & 59.5          & 78.9          & \textbf{63.4}   & 45.4          & \textbf{63.6}  \\
Baseline \cite{15DBLP:journals/corr/abs-2204-14044}                & 64.3          & 52.9          & 38.0          & 51.7          & 73.0          & 58.4          & 41.6          & 57.7          & 77.83         & 61.1  & 43.93         & 61.0           \\
Ours            & \textbf{65.6} & \textbf{55.4} & \textbf{38.8} & \textbf{53.3} & \textbf{75.4} & \textbf{61.0} & \textbf{44.2} & \textbf{60.2} & \textbf{79.8} & 61.4   & \textbf{47.1} & 62.8           \\ 
\hline
C3-STISR-mp \cite{15DBLP:journals/corr/abs-2204-14044}             & 65.2          & 53.6          & 39.8          & 53.7          & 74.2          & 61.0          & 43.2          & 60.5          & 79.1          & 63.3   & 46.8          & 64.1           \\
TPGSR-3\cite{52JianqiMa2021TextPG}                & 63.1          & 52.0          & 38.6          & 51.8          & 74.9          & 60.5          & 44.1          & 60.5          & 78.9          & 62.7   & 44.5          & 62.8           \\
Ours-3         & \textbf{67.0} & \textbf{59.2} & \textbf{42.6} & \textbf{56.3} & \textbf{77.1} & \textbf{63.3} & \textbf{46.5} & \textbf{62.3} & \textbf{80.4} & \textbf{66.1}  & \textbf{49.1} & \textbf{65.2}           \\ 
\bottomrule
\end{tabular}

\end{table*}

\section{Experiments}

In this section, we describe the experiment's implementation details, datasets, and metrics, in Sec.\ref{subsec:setting}. Next, we compare the results of the model with other models, in Sec.\ref{subsec:Comparison}. Then, to verify the generalization of the model, we compare it with other models on the scene text benchmark, in Sec.\ref{4.3}. Finally, we performed ablation experiments to verify the effectiveness of the proposed module, in Sec.\ref{subsec:Ablation}.

\subsection{Experiment Setting}
% \vspace{1 em}
\label{subsec:setting}

\textbf{Implementation Details}. 

Our model is trained on a single Tesla V100 GPU using the Adam optimizer with a training period of 500 epochs and a batch size of 48. The learning rate, beta1, and beta2 are set to 0.001, 0.5, and 0.9, respectively. During the first five epochs, a warm-up strategy is applied and then CosineAnnealingLR is used to adjust the learning rate. The model includes 5 layers of MSRBs. To assess the recognition performance of the super-resolution model, we used common recognizers including CRNN \cite{10]DBLP:journals/pami/ShiBY17}, MORAN \cite{39DBLP:journals/pr/LuoJS19}, and ASTER \cite{38DBLP:journals/pami/ShiYWLYB19}.  The performance of the model is evaluated based on recognition accuracy and image quality, using Peak Signal-to-Noise Ratio (PSNR) and Structural Similarity Index Measure (SSIM) \cite{42DBLP:journals/tip/WangBSS04}. PSNR measures peak error for quick quality assessment, while SSIM considers structural information for more human-perceptual quality assessment. \textbf{Please note that in this article, all the average accuracy values refer to "average accuracy," while the previous work refers to "overall accuracy."} 

\textbf{Scene Text Super-resolution Datasets.} We employed the TextZoom \cite{6]DBLP:conf/eccv/WangX0WLSB20} dataset, a real-world text SR dataset, to train and validate our model. The dataset comprises 21,740 LR-HR text image pairs, with 17,367 samples used for training. These images were captured using cameras with varying focal lengths and were obtained from the SRRAW \cite{40DBLP:journals/corr/abs-1905-05169} and RealSR \cite{41DBLP:journals/corr/abs-1904-00523} image super-resolution datasets. TextZoom is divided into three subsets based on recognition difficulty, namely easy, medium, and hard. The size of the low- and high-resolution image is $64 \times 16$ and $128 \times 32$, respectively.

\textbf{Scene Text Recognition Datasets.} To assess the generalization of our models, we evaluated them on the ICDAR2015 \cite{48DBLP:conf/icdar/KaratzasGNGBIMN15}, COCO text \cite{50veit2016cocotext}, and SVTP \cite{49DBLP:conf/iccv/PhanSTT13} datasets. ICDAR2015 contains 1,811 validation images that were captured using Google Glasses in natural scenes and are plagued by blurring, rotation, and low resolution. SVTP comprises 645 images, mostly featuring curved text and perspective issues. COCO text is a large-scale dataset for text detection and recognition in natural images, and we only used its test set of 9,896 images.

\subsection{Comparison with State-of-the-Arts}
\label{subsec:Comparison}

In this section, we evaluate our model on TextZoom and compare it with existing text image super-resolution models. Our model outperformed the baseline, increasing its performance from 51.7\% to 53.3\% (a 1.6\% improvement). We also compared the recognition accuracy of our model with that of single-stage and multi-stage models. As shown in Tab.\ref{tab:comparison}, we improved the best performance of the one-stage model in CRNN and Moran from 52.6\% and 59.5\% to 53.3\% and 60.2\%, increasing by 0.7\% respectively. With the help of a multi-stage training strategy, we improved the multi-stage SOTA results from 53.7\% to 56.3\%, an improvement of 2.6\%. This demonstrates the effectiveness and advancement of our proposed model.

\begin{figure*}[h!]
  \centering
  \vspace{-1  em}
   %\includegraphics[width=0.8\linewidth]{egfigure.eps}
   \includegraphics[width=0.9\linewidth]{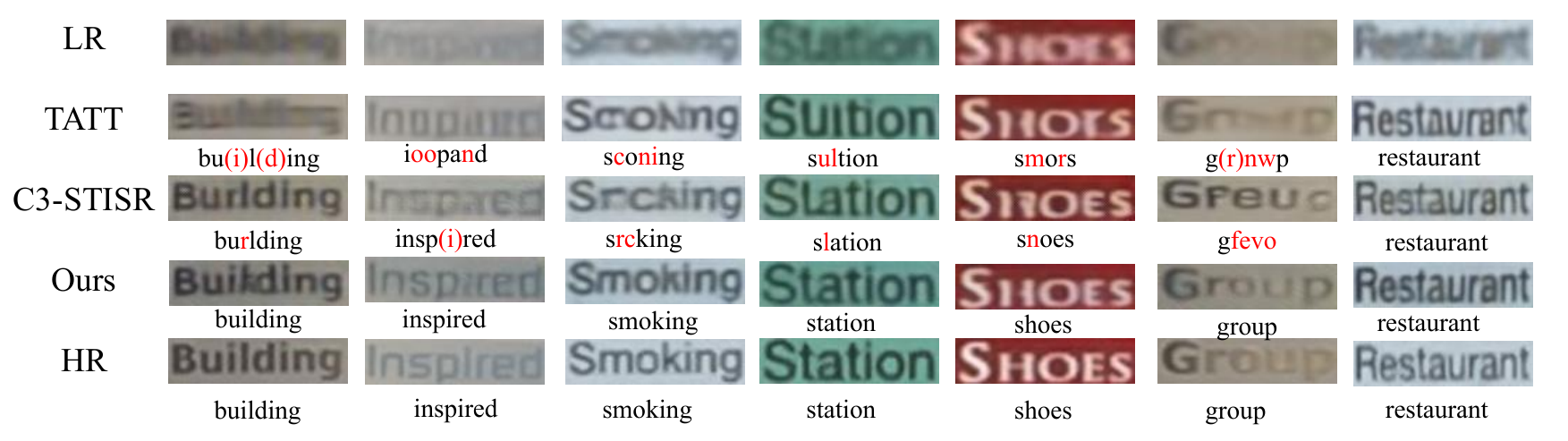}
    \caption{Visualization of super-resolution images and recognition results, red font indicates characters that are misrecognized and the red characters in brackets are undetected characters.}
   \label{fig:6}
\end{figure*}

As shown in Fig.\ref{fig:6}, we visualize the super-resolution image output by the model. Even blurry images due to lighting and deformation are well recovered and recognized by our model. From the perspective of text recovery, the image recovered by our method is more similar to the high-resolution image. The last column in the figure shows that the super-resolution images of each model are obtained and the recognition results are correct. The performance of text restoration in our images is still the most consistent with the ground truth.

\subsection{Result on Scene Text Benchmark}
\label{4.3}
To assess the generalization of the model, we directly tested the model trained on TextZoom on three datasets: ICDAR2015 \cite{48DBLP:conf/icdar/KaratzasGNGBIMN15}, SVTP \cite{49DBLP:conf/iccv/PhanSTT13}, and COCOText \cite{50veit2016cocotext}. These datasets have different labels and data from the TextZoom dataset, which presented a domain shift challenge and provided a robust verification of the generalization of our method. Additionally, we added Gaussian noise to further degrade the image quality, testing the model's ability to generalize under harsh, unpredictable conditions.

\begin{table}[htbp]
\centering
\caption{Performance of our model on Scene Text Benchmark.}
\label{tab:Otherdataset}
% \vspace{0.8  em}
\begin{tabular}{cccc}
\toprule
Input     &ICDAR2015   & SVTP    & COCOText    \\
\midrule
Origin & 64.9      & 69.5    & 40.6   \\
Degrade & 53.2     & 50.7    & 22.2   \\
TATT \cite{12]DBLP:conf/cvpr/MaLZ22}  & 56.4     & 53.3    &25.0    \\
Baseline \cite{15DBLP:journals/corr/abs-2204-14044}  & 59.4    & 55.7    & 26.2   \\
RTSRN & \textbf{61.9}    & \textbf{58.0} & \textbf{28.6} \\
\bottomrule
\end{tabular}

\end{table}

As shown in Tab.\ref{tab:Otherdataset}, the results show that our model outperforms the others. The "Original" and "Degrade" columns represent the original images and the degraded images, respectively. The other columns represent the super-resolution images generated by each model and the recognition accuracy, which was evaluated using the official CRNN. We only used a single-stage model for testing. Our model outperforms others on different datasets and outperforms TATT by 0.6\% when trained and tested on TextZoom. However, in the case of domain shift, our model exceeded TATT by 5.5\%, 4.7\%, and 3.6\% on the three datasets, respectively, demonstrating the generalization ability and robustness of our model.

\subsection{Ablation Study}
\label{subsec:Ablation}

In this section, we will evaluate the effectiveness of each component, including the MSRB, PAM, and $\mathcal{L}_{lca}$. All experiments were validated using CRNN on TextZoom. The metrics used were the average of the results of three subsets, where ACC (\%) represents the accuracy rate and SSIM represents the value in scientific notation (multiplied by $10^{-2}$).

\begin{table}[htb]
\centering
\caption{Vision backbone exploration. The two layers of CNN in SRB are replaced by different single-layer visual extraction blocks.}
\label{tab:type}
% \vspace{0.8  em}
\begin{tabular}{cccc}
\toprule
Type & ACC & PNSR & SSIM\\
\midrule
Baseline           & 51.7     & 20.56  & 74.81  \\
CNN-based SRB           & 51.7     & 20.75  & 74.77  \\
MLP-based SRB          & \textbf{52.8}     & \textbf{20.9}  & \textbf{76.05}  \\ 
Attention-based SRB          & 52.1     & 20.81 & 74.74   \\
\bottomrule
\end{tabular}
\end{table}

\textbf{Vision backbone ablation.} In Tab.\ref{tab:type}, different SRBs mean that the two layers of CNN in SRB \cite{6]DBLP:conf/eccv/WangX0WLSB20} are replaced by ConvNeXt \cite{17DBLP:conf/cvpr/0003MWFDX22}, HaloNet \cite{20DBLP:conf/cvpr/VaswaniRSPHS21} or our proposed MADM feature extraction block. The SRB in Baseline is substituted with these three modules for training and testing. The experimental results show that the visual backbone seriously affects the performance of the model. Under the same experimental setting, SRB is replaced by MSRB, and the recognition accuracy is improved by 1.1\%. MLP-based SRB achieves better results than the other two modules for three reasons. (1) MADM extracts feature in three directions: width, height, and channel, which is more suitable for text images. (2) The dynamic aggregation module proposed by WaveMLP, which regards images as waves, can better aggregate the features along three orthogonal directions to obtain an axial receptive field. (3) The training of the STISR model is similar to the knowledge distillation\cite{43DBLP:journals/corr/HintonVD15}, and the auxiliary model and distillation loss are used. The MLP induction bias is less, and more knowledge can be learned from the auxiliary model.

\begin{table}[htbp]
\centering
\caption{The performance of different operators as refinement modules.}
\label{tab:refine-perfermance}
% \vspace{0.8  em}
\begin{tabular}{cccc}
\toprule
Operator     & ACC  & PNSR    & SSIM    \\
\midrule
Baseline & 51.7     & 20.56   & 74.81  \\
CNN & 51,6    & 20.86   & 75.67  \\
Halo & 51.5    & 19.62   & 75.67  \\
PAM & \textbf{52.3}    & \textbf{20.97} & \textbf{76.28} \\
\bottomrule
\end{tabular}

\end{table}

\textbf{Performance of PAM:} Our proposed PAM improves the feature map after upsampling, which we have verified through comparisons with other modules. We positioned CNN, Halo, and PAM after the upsampling module. As shown in Tab.\ref{tab:refine-perfermance}, the recognition performance of the model dropped when using CNN and Halo as refinement modules. However, our PAM significantly improves the recognition performance and image quality, increasing recognition accuracy from 51.7\% to 52.3\%. This experiment demonstrates that PAM adaptively interacts with the features of center points and neighboring points, capturing structural information and optimizing the details of the feature map, resulting in high-quality super-resolution images.

\begin{figure}
  \centering
   %\includegraphics[width=0.8\linewidth]{egfigure.eps}
   \includegraphics[width=\linewidth]{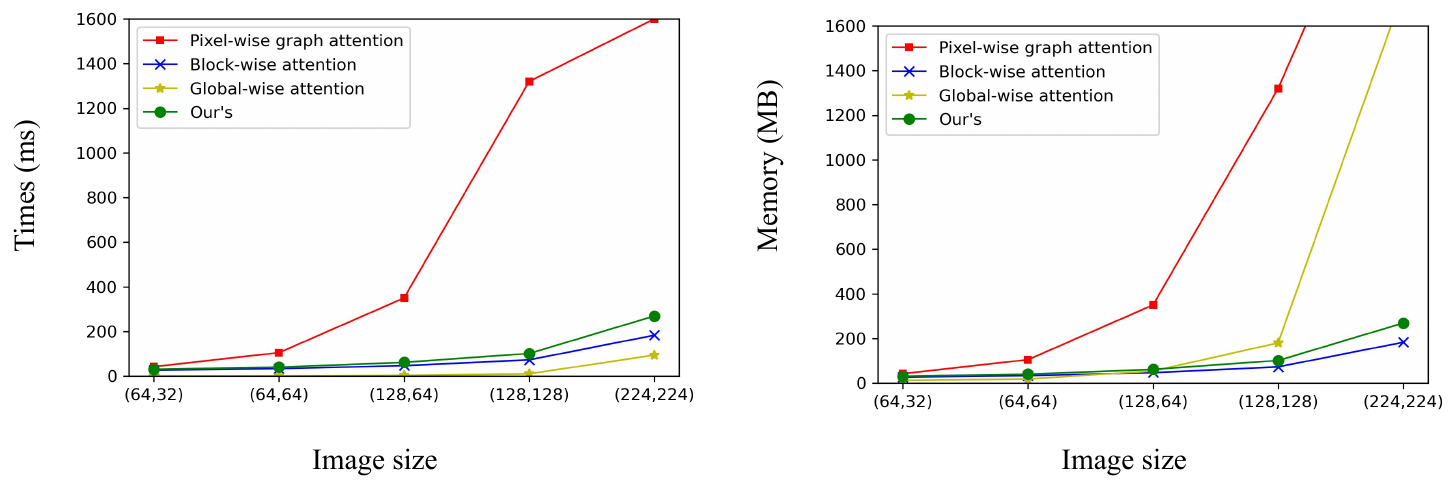}
   \caption{Time and memory cost of different attention modules, like pixel-wise graph attention, block-wise attention, global-wise attention, and our proposed efficient pixel-wise graph attention respectively.}
   \label{fig:time_memory}
\end{figure}

\textbf{Efficiency of PAM.} To verify the efficiency of our proposed efficient pixel-wise graph attention, we compare it with other attention mechanisms. As shown in Fig.\ref{fig:time_memory}, we compare it to pixel-wise graph attention (PGA) \cite{22DBLP:conf/mm/ZhangDHML21}, block-wise attention (halo) \cite{20DBLP:conf/cvpr/VaswaniRSPHS21}, and global-wise attention \cite{59DBLP:conf/nips/VaswaniSPUJGKP17} to calculate their costs. As the image resolution increases, the computational cost of PGA and global attention is much greater than that of Halo and PAM. Because PGA needs to convert pictures to graphs when computing attention. In particular, the graphs converted from pictures are sparse, and the memory usage of GPU is enormous. Although it is also a local attention operator, the cost of computing actually increases dramatically. To solve this problem, we propose to calculate pixel-wise attention using a shift window based on CNN-like methods. \emph{Our improvement reduced the speed by 2-3 orders of magnitude}. The experimental results prove the superiority of our proposed PAM. Halo and PAM are computed in parallel. Therefore, both the GPU memory usage and the time cost are basically consistent with the time complexity analyzed in Sec.\ref{3.3}. PAM is comparable to or even faster than Halo when the resolution is small, and halo is more cost-effective when the resolution increases.

\begin{table}[htbp]
\centering
\caption{Ablation study on the loss function.}
\label{tab:loss-type}
% \vspace{0.8  em}
\begin{tabular}{cccc}
\toprule
Type & ACC & PNSR & SSIM\\
\midrule
 % &    &   &  \\
baseline & 51.7     & 20.56   & 74.81  \\
+$\mathcal{L}_{gp}$ & 51.6    & 20.51   & 73.87  \\
+$\mathcal{L}_{lca}$  & \textbf{52.5}    & \textbf{20.72} & \textbf{76.07} \\
\bottomrule
\end{tabular}

\end{table}

% \usepackage{color}
% \usepackage{tabularray}
% \definecolor{Black}{rgb}{NaN,NaN,NaN}
% \begin{table}
% \centering
% \caption{test}
% \begin{tblr}{
%   cell{2}{1} = {r=3}{},
%   cell{5}{1} = {r=3}{},
%   vline{1} = {1}{},
%   vline{2-3} = {1}{Black},
%   vline{3} = {2}{Black},
%   hline{1,8} = {-}{0.08em},
%   hline{2} = {-}{0.05em},
%   hline{3} = {5}{},
% }
% Type & Factor & ACC           & PNSR           & SSIM           \\
% L1   & 0.01   & 51.6          & \textbf{20.86} & 75.79          \\
%      & 0.1    & \textbf{52.5} & 20.72          & 76.07          \\
%      & 1      & 51.4          & 20.74          & \textbf{76.40} \\
% L2   & 0.01   & 51.6          & 20.83          & 75.66          \\
%      & 0.1    & 52.1          & 20.34          & 74.53          \\
%      & 1      & 51.2          & 18.05          & 67.13    
% \end{tblr}
% \end{table}

\begin{table}[htbp]
\centering
\caption{Influence of calculation method and hyperparameters on results.}
\label{tab:loss-study}
% \vspace{0.8  em}
\begin{tabular}{ccccc}
\toprule
Type & Factor & ACC           & PNSR           & SSIM           \\
\midrule
\multirow{3}*{L1}   & 0.01   & 51.6          & \textbf{20.86} & 75.79          \\
 ~   & 0.1    & \textbf{52.5} & 20.72          & 76.07          \\
 ~   & 1      & 51.4          & 20.74          & \textbf{76.40} \\
 \hline
\multirow{3}*{L2}   & 0.01   & 51.6          & 20.83          & 75.66          \\
 ~   & 0.1    & 52.1          & 20.34          & 74.53          \\
 ~    & 1      & 51.2          & 18.05          & 67.13    \\
\bottomrule
\end{tabular}

\end{table}

\textbf{Loss function ablation.} We compare our loss function with gradient profile loss \cite{6]DBLP:conf/eccv/WangX0WLSB20} which is also based on gradient. We use $\mathcal{L}_{gp}$ to represent Gradient Profile Loss. The results are shown in Tab.\ref{tab:loss-type}. Our loss performs better in recognition results and image quality. This is because $\mathcal{L}_{gp}$ calculates the gradient profile of the whole image, and it is difficult to obtain fine local details. The local contour awareness loss $\mathcal{L}_{lca}$ based on HOG \cite{33DBLP:conf/mmm/HuangTHTJ11} operates on the local grid cell of the image, preserving good invariance to the image's geometric and optical deformations.

In order to optimize the use of $\mathcal{L}_{lca}$, we investigated the calculation methods from [L1, L2] and various hyperparameters from [0.01, 0.1, 1]. It's worth mentioning that the hyperparameters for other loss functions were kept at the default values specified in the baseline \cite{15DBLP:journals/corr/abs-2204-14044}. The results are depicted in Tab.\ref{tab:loss-study}. The combination of L1 and 0.1 produced the best performance. The reason is that the HOG features are already regularized, so if another L2 regularization is applied, the loss becomes smoothed and the model becomes difficult to optimize.

\begin{table}
\small
\centering
\caption{Ablation experiments in different model architectures.}
\label{tab:architecture-ablation}
\begin{tabular}{cccccccc} 
\toprule
MSRB & PAM & $\mathcal{L}_{lca}$ & Parameters  & FLOPs & ACC & PNSR    & SSIM    \\ 
\midrule
    &     &     & 17.15 M	  & 2.52 GMac & 51.7     & 20.56 & 74.81  \\ 

  \color{red} \checkmark   &     &  & 17.03 M	  & 2.41 GMac    & 52.8     & 20.85 & 76.05  \\
    &   \color{red} \checkmark   &   & 17.16 M	  & 2.57 GMac   & 52.3    & 20.97 & 76.28  \\
    &     &   \color{red} \checkmark  & 17.15 M	  & 2.52 GMac  & 52.5    & 20.72 & 76.07  \\
    \midrule
  \color{red} \checkmark   &   \color{red} \checkmark   &  & 17.04 M	  & 2.46 GMac    & 52.7    & 19.44 & 75.48  \\
  \color{red} \checkmark   &     &   \color{red} \checkmark & 17.03 M	  & 2.41 GMac   & 52.9    & 19.11 & 75.2   \\
    &   \color{red} \checkmark   &   \color{red} \checkmark   & 17.16 M	  & 2.57 GMac & 52.3    & 19.73 & 75.4   \\ 
\midrule
  \color{red} \checkmark   &   \color{red} \checkmark   &   \color{red} \checkmark & 17.04 M	  & 2.46 GMac  & 53.3    & 20.16 & 76.19  \\
\bottomrule
\end{tabular}

\end{table}

\textbf{Architecture ablation.} As shown in Tab.\ref{tab:architecture-ablation}, our experimental results show the effectiveness of our module. The data presented in the first four rows of the table indicate a noticeable improvement in the performance of the model as the corresponding module is added. In addition, the results in the seventh row of the table show an interesting phenomenon. If we do not replace them with a new vision backbone, the results are not even as good as when using each module alone. This result corroborates our view that how to better extract image features is the basis for super-resolution tasks. Our proposed module can extract robust features from low-quality images for upsampling, and the post-processing module effectively mitigates the degradation of image quality caused by upsampling.

\section{Conclusion}
In this paper, we propose a post-processing module Pixel Adapter to alleviate the issue of image pixel distortion caused by upsampling. This module is based on our newly introduced efficient pixel-wise graph attention, significantly improving runtime efficiency and substantially reducing memory usage. Additionally, we combine more advanced feature processing modules MSRB and a loss function $\mathcal{L}_{lca}$ that emphasizes local details, resulting in a lightweight text image super-resolution framework. In the future, we will validate the effectiveness of our proposed method in general super-resolution, image denoising and restoration tasks.

\begin{acks}
This work was supported by the National Key Research and Development Program of China under Grant 2020YFB2103803. This research was supported by the Supercomputing Center of University of Science and Technology of China.
\end{acks}
\bibliographystyle{unsrt}
\balance

\bibliography{sample-base}

\begin{thebibliography}{10}

\bibitem{6]DBLP:conf/eccv/WangX0WLSB20}
Wenjia Wang, Enze Xie, Xuebo Liu, Wenhai Wang, Ding Liang, Chunhua Shen, and Xiang Bai.
\newblock Scene text image super-resolution in the wild.
\newblock In {\em Computer Vision - {ECCV} 2020 - 16th European Conference, Glasgow, UK, August 23-28, 2020, Proceedings, Part {X}}, volume 12355 of {\em Lecture Notes in Computer Science}, pages 650--666. Springer, 2020.

\bibitem{3]DBLP:journals/eswa/KhareSCLLHB19}
Vijeta Khare, Palaiahnakote Shivakumara, Chee~Seng Chan, Tong Lu, Kim~Meng Liang, Hock~Woon Hon, and Michael Blumenstein.
\newblock A novel character segmentation-reconstruction approach for license plate recognition.
\newblock {\em Expert Syst. Appl.}, 131:219--239, 2019.

\bibitem{1]DBLP:journals/cviu/FangFYCC04}
Chiung{-}Yao Fang, Chiou{-}Shann Fuh, P.~S. Yen, Shen Cherng, and Sei{-}Wang Chen.
\newblock An automatic road sign recognition system based on a computational model of human recognition processing.
\newblock {\em Comput. Vis. Image Underst.}, 96(2):237--268, 2004.

\bibitem{2]DBLP:journals/tits/ZhangDPFW21}
Chongsheng Zhang, Weiping Ding, Guowen Peng, Feifei Fu, and Wei Wang.
\newblock Street view text recognition with deep learning for urban scene understanding in intelligent transportation systems.
\newblock {\em {IEEE} Trans. Intell. Transp. Syst.}, 22(7):4727--4743, 2021.

\bibitem{4]20190906545388}
Hoang~Danh Liem, Nguyen~Duc Minh, Nguyen~Bao Trung, Hoang~Tien Duc, Pham~Hoang Hiep, Doan~Viet Dung, and Dang~Hoang Vu.
\newblock Fvi: An end-to-end vietnamese identification card detection and recognition in images.
\newblock pages 338 -- 340, 2018.

\bibitem{5]DBLP:journals/pr/SanchezRTVV19}
Joan{-}Andreu S{\'{a}}nchez, Ver{\'{o}}nica Romero, Alejandro~H. Toselli, Mauricio Villegas, and Enrique Vidal.
\newblock A set of benchmarks for handwritten text recognition on historical documents.
\newblock {\em Pattern Recognit.}, 94:122--134, 2019.

\bibitem{12]DBLP:conf/cvpr/MaLZ22}
Jianqi Ma, Zhetong Liang, and Lei Zhang.
\newblock A text attention network for spatial deformation robust scene text image super-resolution.
\newblock In {\em {IEEE/CVF} Conference on Computer Vision and Pattern Recognition, {CVPR} 2022, New Orleans, LA, USA, June 18-24, 2022}, pages 5901--5910. {IEEE}, 2022.

\bibitem{15DBLP:journals/corr/abs-2204-14044}
Minyi Zhao, Miao Wang, Fan Bai, Bingjia Li, Jie Wang, and Shuigeng Zhou.
\newblock {C3-STISR:} scene text image super-resolution with triple clues.
\newblock {\em CoRR}, abs/2204.14044, 2022.

\bibitem{13DBLP:conf/cvpr/ChenLX21}
Jingye Chen, Bin Li, and Xiangyang Xue.
\newblock Scene text telescope: Text-focused scene image super-resolution.
\newblock In {\em {IEEE} Conference on Computer Vision and Pattern Recognition, {CVPR} 2021, virtual, June 19-25, 2021}, pages 12026--12035. Computer Vision Foundation / {IEEE}, 2021.

\bibitem{11]DBLP:journals/corr/abs-2106-15368}
Jianqi Ma, Shi Guo, and Lei Zhang.
\newblock Text prior guided scene text image super-resolution.
\newblock {\em CoRR}, abs/2106.15368, 2021.

\bibitem{55DBLP:conf/eccv/DongLHT14}
Chao Dong, Chen~Change Loy, Kaiming He, and Xiaoou Tang.
\newblock Learning a deep convolutional network for image super-resolution.
\newblock In David~J. Fleet, Tom{\'{a}}s Pajdla, Bernt Schiele, and Tinne Tuytelaars, editors, {\em Computer Vision - {ECCV} 2014 - 13th European Conference, Zurich, Switzerland, September 6-12, 2014, Proceedings, Part {IV}}, volume 8692 of {\em Lecture Notes in Computer Science}, pages 184--199. Springer, 2014.

\bibitem{56DBLP:conf/cvpr/ShiCHTABRW16}
Wenzhe Shi, Jose Caballero, Ferenc Huszar, Johannes Totz, Andrew~P. Aitken, Rob Bishop, Daniel Rueckert, and Zehan Wang.
\newblock Real-time single image and video super-resolution using an efficient sub-pixel convolutional neural network.
\newblock In {\em 2016 {IEEE} Conference on Computer Vision and Pattern Recognition, {CVPR} 2016, Las Vegas, NV, USA, June 27-30, 2016}, pages 1874--1883. {IEEE} Computer Society, 2016.

\bibitem{57DBLP:journals/corr/abs-2210-05171}
Man Zhou, Hu~Yu, Jie Huang, Feng Zhao, Jinwei Gu, Chen~Change Loy, Deyu Meng, and Chongyi Li.
\newblock Deep fourier up-sampling.
\newblock {\em CoRR}, abs/2210.05171, 2022.

\bibitem{58DBLP:conf/iclr/VelickovicCCRLB18}
Petar Velickovic, Guillem Cucurull, Arantxa Casanova, Adriana Romero, Pietro Li{\`{o}}, and Yoshua Bengio.
\newblock Graph attention networks.
\newblock In {\em 6th International Conference on Learning Representations, {ICLR} 2018, Vancouver, BC, Canada, April 30 - May 3, 2018, Conference Track Proceedings}, 2018.

\bibitem{59DBLP:conf/nips/VaswaniSPUJGKP17}
Ashish Vaswani, Noam Shazeer, Niki Parmar, Jakob Uszkoreit, Llion Jones, Aidan~N. Gomez, Lukasz Kaiser, and Illia Polosukhin.
\newblock Attention is all you need.
\newblock In {\em Advances in Neural Information Processing Systems 30: Annual Conference on Neural Information Processing Systems 2017, December 4-9, 2017, Long Beach, CA, {USA}}, pages 5998--6008, 2017.

\bibitem{20DBLP:conf/cvpr/VaswaniRSPHS21}
Ashish Vaswani, Prajit Ramachandran, Aravind Srinivas, Niki Parmar, Blake~A. Hechtman, and Jonathon Shlens.
\newblock Scaling local self-attention for parameter efficient visual backbones.
\newblock In {\em {IEEE} Conference on Computer Vision and Pattern Recognition, {CVPR} 2021, virtual, June 19-25, 2021}, pages 12894--12904. Computer Vision Foundation / {IEEE}, 2021.

\bibitem{22DBLP:conf/mm/ZhangDHML21}
Wenyu Zhang, Qing Ding, Jian Hu, Yi~Ma, and Mingzhe Lu.
\newblock Pixel-wise graph attention networks for person re-identification.
\newblock In {\em {MM} '21: {ACM} Multimedia Conference, Virtual Event, China, October 20 - 24, 2021}, pages 5231--5238. {ACM}, 2021.

\bibitem{60DBLP:conf/cvpr/DalalT05}
Navneet Dalal and Bill Triggs.
\newblock Histograms of oriented gradients for human detection.
\newblock In {\em 2005 {IEEE} Computer Society Conference on Computer Vision and Pattern Recognition {(CVPR} 2005), 20-26 June 2005, San Diego, CA, {USA}}, pages 886--893. {IEEE} Computer Society, 2005.

\bibitem{28DBLP:conf/codit/FerjaouiCAZ22}
Radhia Ferjaoui, Mohamed~Ali Cherni, Fathia Abidi, and Asma Zidi.
\newblock Deep residual learning based on resnet50 for {COVID-19} recognition in lung {CT} images.
\newblock In {\em 8th International Conference on Control, Decision and Information Technologies, CoDIT 2022, Istanbul, Turkey, May 17-20, 2022}, pages 407--412. {IEEE}, 2022.

\bibitem{21DBLP:journals/corr/abs-2107-00782}
Huajun Liu, Fuqiang Liu, Xinyi Fan, and Dong Huang.
\newblock Polarized self-attention: Towards high-quality pixel-wise regression.
\newblock {\em CoRR}, abs/2107.00782, 2021.

\bibitem{25DBLP:conf/nips/ParmarRVBLS19}
Niki Parmar, Prajit Ramachandran, Ashish Vaswani, Irwan Bello, Anselm Levskaya, and Jonathon Shlens.
\newblock Stand-alone self-attention in vision models.
\newblock In {\em Advances in Neural Information Processing Systems 32: Annual Conference on Neural Information Processing Systems 2019, NeurIPS 2019, December 8-14, 2019, Vancouver, BC, Canada}, pages 68--80, 2019.

\bibitem{27DBLP:conf/iclr/DosovitskiyB0WZ21}
Alexey Dosovitskiy, Lucas Beyer, Alexander Kolesnikov, Dirk Weissenborn, Xiaohua Zhai, Thomas Unterthiner, Mostafa Dehghani, Matthias Minderer, Georg Heigold, Sylvain Gelly, Jakob Uszkoreit, and Neil Houlsby.
\newblock An image is worth 16x16 words: Transformers for image recognition at scale.
\newblock In {\em 9th International Conference on Learning Representations, {ICLR} 2021, Virtual Event, Austria, May 3-7, 2021}. OpenReview.net, 2021.

\bibitem{1920210401013}
Yehui Tang, Kai Han, Jianyuan Guo, Chang Xu, Yanxi Li, Chao Xu, and Yunhe Wang.
\newblock An image patch is a wave: Quantum inspired vision mlp.
\newblock 2021.

\bibitem{18DBLP:journals/corr/abs-2105-01601}
Ilya~O. Tolstikhin, Neil Houlsby, Alexander Kolesnikov, Lucas Beyer, Xiaohua Zhai, Thomas Unterthiner, Jessica Yung, Andreas Steiner, Daniel Keysers, Jakob Uszkoreit, Mario Lucic, and Alexey Dosovitskiy.
\newblock Mlp-mixer: An all-mlp architecture for vision.
\newblock {\em CoRR}, abs/2105.01601, 2021.

\bibitem{9]DBLP:conf/rivf/TranH19}
Hanh T.~M. Tran and Tien Ho{-}Phuoc.
\newblock Deep laplacian pyramid network for text images super-resolution.
\newblock In {\em 2019 {IEEE-RIVF} International Conference on Computing and Communication Technologies, {RIVF} 2019, Danang, Vietnam, March 20-22, 2019}, pages 1--6. {IEEE}, 2019.

\bibitem{10]DBLP:journals/pami/ShiBY17}
Baoguang Shi, Xiang Bai, and Cong Yao.
\newblock An end-to-end trainable neural network for image-based sequence recognition and its application to scene text recognition.
\newblock {\em {IEEE} Trans. Pattern Anal. Mach. Intell.}, 39(11):2298--2304, 2017.

\bibitem{51DBLP:conf/cvpr/LedigTHCCAATTWS17}
Christian Ledig, Lucas Theis, Ferenc Huszar, Jose Caballero, Andrew Cunningham, Alejandro Acosta, Andrew~P. Aitken, Alykhan Tejani, Johannes Totz, Zehan Wang, and Wenzhe Shi.
\newblock Photo-realistic single image super-resolution using a generative adversarial network.
\newblock In {\em 2017 {IEEE} Conference on Computer Vision and Pattern Recognition, {CVPR} 2017, Honolulu, HI, USA, July 21-26, 2017}, pages 105--114. {IEEE} Computer Society, 2017.

\bibitem{52JianqiMa2021TextPG}
Jianqi Ma, Shi Guo, and Lei Zhang.
\newblock Text prior guided scene text image super-resolution.
\newblock {\em arXiv: Computer Vision and Pattern Recognition}, 2021.

\bibitem{30DBLP:journals/corr/abs-1801-04381}
Mark Sandler, Andrew~G. Howard, Menglong Zhu, Andrey Zhmoginov, and Liang{-}Chieh Chen.
\newblock Inverted residuals and linear bottlenecks: Mobile networks for classification, detection and segmentation.
\newblock {\em CoRR}, abs/1801.04381, 2018.

\bibitem{14DBLP:journals/corr/abs-2112-08171}
Jingye Chen, Haiyang Yu, Jianqi Ma, Bin Li, and Xiangyang Xue.
\newblock Text gestalt: Stroke-aware scene text image super-resolution.
\newblock {\em CoRR}, abs/2112.08171, 2021.

\bibitem{53DBLP:conf/cvpr/GuptaVZ16}
Ankush Gupta, Andrea Vedaldi, and Andrew Zisserman.
\newblock Synthetic data for text localisation in natural images.
\newblock In {\em 2016 {IEEE} Conference on Computer Vision and Pattern Recognition, {CVPR} 2016, Las Vegas, NV, USA, June 27-30, 2016}, pages 2315--2324. {IEEE} Computer Society, 2016.

\bibitem{54DBLP:journals/ijcv/JaderbergSVZ16}
Max Jaderberg, Karen Simonyan, Andrea Vedaldi, and Andrew Zisserman.
\newblock Reading text in the wild with convolutional neural networks.
\newblock {\em Int. J. Comput. Vis.}, 116(1):1--20, 2016.

\bibitem{7]DBLP:conf/nips/JaderbergSZK15}
Max Jaderberg, Karen Simonyan, Andrew Zisserman, and Koray Kavukcuoglu.
\newblock Spatial transformer networks.
\newblock In {\em Advances in Neural Information Processing Systems 28: Annual Conference on Neural Information Processing Systems 2015, December 7-12, 2015, Montreal, Quebec, Canada}, pages 2017--2025, 2015.

\bibitem{17DBLP:conf/cvpr/0003MWFDX22}
Zhuang Liu, Hanzi Mao, Chao{-}Yuan Wu, Christoph Feichtenhofer, Trevor Darrell, and Saining Xie.
\newblock A convnet for the 2020s.
\newblock In {\em {IEEE/CVF} Conference on Computer Vision and Pattern Recognition, {CVPR} 2022, New Orleans, LA, USA, June 18-24, 2022}, pages 11966--11976. {IEEE}, 2022.

\bibitem{23DBLP:journals/corr/abs-1710-10903}
Petar Velickovic, Guillem Cucurull, Arantxa Casanova, Adriana Romero, Pietro Li{\`{o}}, and Yoshua Bengio.
\newblock Graph attention networks.
\newblock {\em CoRR}, abs/1710.10903, 2017.

\bibitem{39DBLP:journals/pr/LuoJS19}
Canjie Luo, Lianwen Jin, and Zenghui Sun.
\newblock {MORAN:} {A} multi-object rectified attention network for scene text recognition.
\newblock {\em Pattern Recognit.}, 90:109--118, 2019.

\bibitem{38DBLP:journals/pami/ShiYWLYB19}
Baoguang Shi, Mingkun Yang, Xinggang Wang, Pengyuan Lyu, Cong Yao, and Xiang Bai.
\newblock {ASTER:} an attentional scene text recognizer with flexible rectification.
\newblock {\em {IEEE} Trans. Pattern Anal. Mach. Intell.}, 41(9):2035--2048, 2019.

\bibitem{42DBLP:journals/tip/WangBSS04}
Zhou Wang, Alan~C. Bovik, Hamid~R. Sheikh, and Eero~P. Simoncelli.
\newblock Image quality assessment: from error visibility to structural similarity.
\newblock {\em {IEEE} Trans. Image Process.}, 13(4):600--612, 2004.

\bibitem{40DBLP:journals/corr/abs-1905-05169}
Xuaner~Cecilia Zhang, Qifeng Chen, Ren Ng, and Vladlen Koltun.
\newblock Zoom to learn, learn to zoom.
\newblock {\em CoRR}, abs/1905.05169, 2019.

\bibitem{41DBLP:journals/corr/abs-1904-00523}
Jianrui Cai, Hui Zeng, Hongwei Yong, Zisheng Cao, and Lei Zhang.
\newblock Toward real-world single image super-resolution: {A} new benchmark and {A} new model.
\newblock {\em CoRR}, abs/1904.00523, 2019.

\bibitem{48DBLP:conf/icdar/KaratzasGNGBIMN15}
Dimosthenis Karatzas, Lluis Gomez{-}Bigorda, Anguelos Nicolaou, Suman~K. Ghosh, Andrew~D. Bagdanov, Masakazu Iwamura, Jiri Matas, Lukas Neumann, Vijay~Ramaseshan Chandrasekhar, Shijian Lu, Faisal Shafait, Seiichi Uchida, and Ernest Valveny.
\newblock {ICDAR} 2015 competition on robust reading.
\newblock In {\em 13th International Conference on Document Analysis and Recognition, {ICDAR} 2015, Nancy, France, August 23-26, 2015}, pages 1156--1160. {IEEE} Computer Society, 2015.

\bibitem{50veit2016cocotext}
Andreas Veit, Tomas Matera, Lukas Neumann, Jiri Matas, and Serge Belongie.
\newblock Coco-text: Dataset and benchmark for text detection and recognition in natural images.
\newblock In {\em arXiv preprint arXiv:1601.07140}, 2016.

\bibitem{49DBLP:conf/iccv/PhanSTT13}
Trung~Quy Phan, Palaiahnakote Shivakumara, Shangxuan Tian, and Chew~Lim Tan.
\newblock Recognizing text with perspective distortion in natural scenes.
\newblock In {\em {IEEE} International Conference on Computer Vision, {ICCV} 2013, Sydney, Australia, December 1-8, 2013}, pages 569--576. {IEEE} Computer Society, 2013.

\bibitem{43DBLP:journals/corr/HintonVD15}
Geoffrey~E. Hinton, Oriol Vinyals, and Jeffrey Dean.
\newblock Distilling the knowledge in a neural network.
\newblock {\em CoRR}, abs/1503.02531, 2015.

\bibitem{29DBLP:journals/tip/SunSXS11}
Jian Sun, Jian Sun, Zongben Xu, and Heung{-}Yeung Shum.
\newblock Gradient profile prior and its applications in image super-resolution and enhancement.
\newblock {\em {IEEE} Trans. Image Process.}, 20(6):1529--1542, 2011.

\bibitem{35DBLP:journals/corr/abs-2208-07818}
Yang Zhi{-}Han.
\newblock Training latent variable models with auto-encoding variational bayes: {A} tutorial.
\newblock {\em CoRR}, abs/2208.07818, 2022.

\bibitem{31DBLP:conf/iccv/LiuL00W0LG21}
Ze~Liu, Yutong Lin, Yue Cao, Han Hu, Yixuan Wei, Zheng Zhang, Stephen Lin, and Baining Guo.
\newblock Swin transformer: Hierarchical vision transformer using shifted windows.
\newblock In {\em 2021 {IEEE/CVF} International Conference on Computer Vision, {ICCV} 2021, Montreal, QC, Canada, October 10-17, 2021}, pages 9992--10002. {IEEE}, 2021.

\end{thebibliography}
% \printbibliography

\clearpage
\appendix

\section{Text Information Extraction}
\label{appdenx.1}

\begin{figure*}
  \centering
   %\includegraphics[width=0.8\linewidth]{egfigure.eps}
   \includegraphics[width=\linewidth]{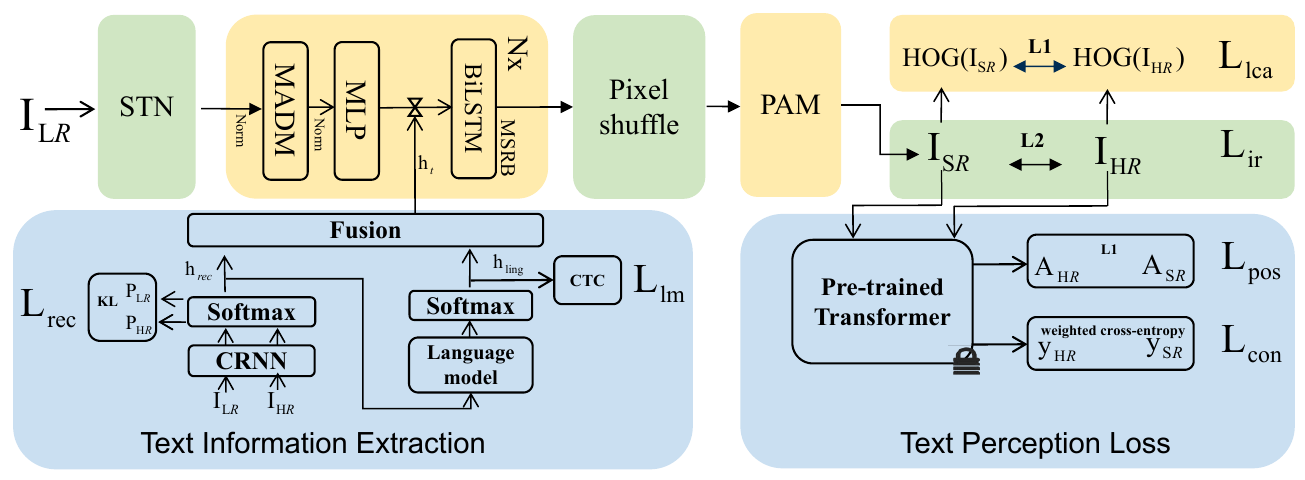}
   \caption{Text Information Extraction and all loss functions used in training.}
   \label{fig:appendix1}
\end{figure*}

For text information extraction, as shown in Fig,\ref{fig:appendix1}, the low-resolution image $I_{LR}$ is input into the text information extraction path to extract text clues. The text probability $h_{rec}$ of the text in the image is obtained using a CRNN \cite{10]DBLP:journals/pami/ShiBY17}. However, recognition models have limitations in terms of modal compatibility and recognition accuracy. Therefore, following C3-STISR \cite{15DBLP:journals/corr/abs-2204-14044}, we used a pre-trained character-level language model to correct the predicted text and obtain linguistic cues $h_{ling}$. These text clues are then fused in the fusion module to form the final text clues  $h_t$, which are transformed into a unified pixel feature map of $C \times H \times W$.

\section{Loss Function in this paper}
\label{appdenx.2}

Blurred text images are often caused by factors such as poor lighting, rotation, and out-of-focus. Loss functions based on image priors, such as L1 or L2 loss, are commonly used to supervise the reconstruction of image pixels or gradients in previous works. For example, Wang et al. \cite{6]DBLP:conf/eccv/WangX0WLSB20} utilized the Gradient Profile Prior \cite{29DBLP:journals/tip/SunSXS11} for image super-resolution tasks by using the gradient profile of the image as the reconstruction target. However, this approach does not effectively address the problems caused by insufficient lighting and focal length disorders. To overcome these limitations, we propose a local contour awareness loss based on image priors and utilize the Histogram of Oriented Gradients (HOG) \cite{33DBLP:conf/mmm/HuangTHTJ11} for the super-resolution task of images. HOG is advantageous because it can effectively capture local shapes by extracting edge and gradient features, and has good invariance to geometric and optical changes due to local extraction of the image through gamma correction and gradient direction quantization. The proposed loss is as follows:
\begin{equation}
\mathcal{L}_{\mathrm{lca}}=\left\|\mathbf{HOG}(\mathbf{I}_{\mathrm{HR}})-\mathbf{HOG}(\mathbf{I}_{\mathrm{SR}})\right\|_1,
\end{equation}

In our work, given a pair of high-resolution (HR) and super-resolution (SR) images $(\mathbf{I}_{\mathrm{HR}},\mathbf{I}_{\mathrm{SR}})$, we aim to minimize the difference in their HOG features. The computation of HOG involves the following steps: Firstly, gradient information is extracted from the image by applying gradient operators. These gradient values provide information about the intensity changes in the image and direction information about the edges in the image. Secondly, the gradient information is quantized into a set of predetermined orientations, typically defined in terms of a histogram of directions. Thirdly, the quantized gradient information is divided into a set of non-overlapping cells, and a histogram is computed for each cell. Finally, these local histograms are concatenated to form a feature descriptor that characterizes the structural information in the image. These details are further explained in the appendix \ref{appdenx.4}. The HOG operator operates on the local square units of the image, ensuring robustness against geometric and optical changes. To further improve image reconstruction, we also incorporate an L2 pixel loss, $\mathcal{L}_{p i x}=\left\|I_{H R}-I_{S R}\right\|_2$, into the image reconstruction loss, i.e.,

\begin{equation}
\mathcal{L}_{\mathrm{ir}}=\mathcal{L}_{p i x} + \mathcal{L}_{\mathrm{lca}},
\end{equation}

 Referring to STT \cite{13DBLP:conf/cvpr/ChenLX21}, we use a pre-trained transformer to obtain a series of attention maps represented as $\mathbf{A} = [a_1,a_2,...a_l]$ and character level predictions, represented as $\mathbf{O} = [o_1,o_2,...o_l]$. The attention maps serve as the label supervision model during the training of character regions, with the content-aware loss $\mathcal{L}_{\text {pos}}=\left\|\mathbf{A}_{\mathrm{HR}}-\mathbf{A}_{\mathrm{SR}}\right\|_1$. Content-Aware loss alleviates characters that are easily confused. A variational automatic encoder is used to obtain the potential features of characters and then calculate the confusion weight $c_{ij}$. When $i=j$, we set $c_{ij}=1$. Combined character prediction results, weighted activation is $a_j=\frac{\mathrm{e}^{o_j}}{\sum_{i=1}^{|\mathcal{A}|} c_{i j} \mathrm{e}^{o_i}}$, where $\mathcal{A}$ represents the alphabet, $o_j$ represents the predicted outcome, and $a_j$ is the weight activation of the j-th character. Content-Aware loss is expressed as $\mathcal{L}_{\mathrm{con}}=-\sum_t \ln a_{y t}$, where $yt$ denotes the ground truth at the t-th time step. 

The text perception loss is as follows:
\begin{equation}
\mathcal{L}_{\mathrm{tp}}=\mathcal{L}_{pos} + \mathcal{L}_{\mathrm{con}},
\end{equation}

This paper employs text clues in the super-resolution process, where the accuracy of the recognizer is critical to the overall performance of the model. To improve the performance of the recognizer, it needs to be optimized in the training stage. The recognizer is first used to obtain the text probability $\mathbf{P_{lr}} \in \mathbb{R}^{L \times|\mathcal{A}|}$ of the low-resolution text image $\mathbf{I}_{LR}$, where $|\mathcal{A}|$ is the length of the alphabet and $\mathbf{L}$ represents the length of the text. To optimize the recognizer, we adopt the distillation loss method \cite{43DBLP:journals/corr/HintonVD15} and express the loss as $\mathcal{L}_{\mathrm{rec}} = \mathbf{KL}(\mathbf{P_{lr}}, \mathbf{P}_{hr})$. The C3-STISR then corrects $\mathbf{P_{lr}}$ through the language model $f_{LM}$ to produce the corrected probability distribution $p_{L M}$, which is computed as $\mathbf{p}_{L M}=f_{L M}\left(R\left(\mathbf{I_{lr}}\right)\right)$. Finally, the CTC loss \cite{35DBLP:journals/corr/abs-2208-07818} is used to align the language model's revised results with the ground truth, expressed as $\mathcal{L}_{\mathrm{lm}} = \mathbf{CTC}(\mathbf{P_{L M}}, \mathbf{target})$.

The text recognition losses are as follows:
\begin{equation}
\mathcal{L}_{\mathrm{tr}}=\mathcal{L}_{\mathrm{rec}} + \mathcal{L}_{\mathrm{lm}},
\end{equation}

The total loss is as follows:
\begin{equation}
\mathcal{L}=\mathcal{L}_{\mathrm{ir}} + \mathcal{L}_{\mathrm{tp}}+ \mathcal{L}_{\mathrm{tr}},
\end{equation}

\section{Vision Backbone exploration}
\label{appdenx.3}

\begin{figure}
  \centering
   %\includegraphics[width=0.8\linewidth]{egfigure.eps}
   \includegraphics[width=\linewidth]{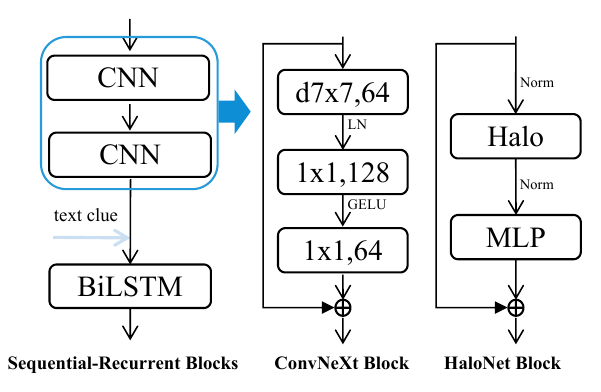}
   \caption{Vision Backbone exploration: exploring the impact of different feature extraction operators on performance.}
   \label{fig:appendix2}
\end{figure}

In the field of scene text image super-resolution, the traditional approach has been to employ Sequential Recurrent Blocks (SRBs) \cite{6]DBLP:conf/eccv/WangX0WLSB20} for feature extraction. However, the limitations of the simple two-layer CNN used in SRBs have become increasingly apparent in meeting the demands of this task. As shown in Fig.\ref{fig:appendix2}, we have explored the replacement of the two-layer CNNs in SRBs with more sophisticated operators, including CNNs, MLPs, and Attention. The features acquired by our feature extraction block will play a crucial role in the subsequent super-resolution process and greatly improve the performance of the method.

\subsection{CNN based block}
ConvNeXt \cite{17DBLP:conf/cvpr/0003MWFDX22} has received much attention for its faster inference speed and higher accuracy than Swin Transformer \cite{31DBLP:conf/iccv/LiuL00W0LG21} under the same FLOPs using a convolutional structure. There is no particularly complex or innovative structure, it borrows ideas from Swin Transformer from the five perspectives of macro design, depthwise convolution, inverted bottleneck \cite{30DBLP:journals/corr/abs-1801-04381}, large kernel, and micro design. As shown in \ref{fig:appendix2}, the ConvNeXt block uses the structure of the inverted bottleneck to replace the regular CNN with depthwise convolution, achieving a better balance between FLOPs and accuracy. At the same time, Depthwise Convolution advances and expands the convolution kernel. The activation function and normalization in each block are reduced. and replaced with GELU and Layer Normalization (LN), respectively.

\subsection{Attention based block}
HaloNet \cite{20DBLP:conf/cvpr/VaswaniRSPHS21} are designed to capture the relationships between different regions of the input image, allowing the network to focus on the most important parts of the image.In practice, the self-attention mechanism is implemented by using two separate operations: a query operation and a key-value operation. The query operation generates a representation of the current position in the image, while the key-value operation generates a representation of all other positions in the image. The attention weights are then computed as the dot product between the query and key representations and are used to scale the corresponding value representation. As shown in Fig.\ref{fig:appendix2}, the resulting self-attention maps are then concatenated with the original feature maps and passed through a feedforward network to produce the final output

\section{HOG details}
\label{appdenx.4}

Hog is a way to manually extract image features. It counts the gradient direction information of the local area of the image as the feature of the area. Firstly, we use a convolution-like operation to obtain the small sum of waiting for directions for generating gradients (subtracting adjacent pixels) on the x and y axes and obtain the phase in combination with the arctan function. Secondly, divide the gradient into different cells, and accumulate the gradient into several bin-direction histogram vectors. Finally, the histogram was obtained by standardization. So we get the HOG feature. More importantly, this algorithm can obtain HOG features efficiently and conveniently, with negligible computational overhead.

Let an image be represented by $I$, and let $C$ be the set of cells dividing the image into non-overlapping regions. The gradient direction within each cell $c \in C$ is calculated as:

\begin{equation}
    g(x,y) = \arctan\left(\frac{\partial I}{\partial y}(x,y) \Big/ \frac{\partial I}{\partial x}(x,y)\right),
\end{equation}

where $\frac{\partial I}{\partial x}$ and $\frac{\partial I}{\partial y}$ are the partial derivatives of the image with respect to the x and y axes, respectively.

Next, a histogram of gradient directions is calculated for each cell. Let $B$ be the set of bins used to construct the histogram, and let $b \in B$ be a bin representing a range of gradient directions. The value of the histogram for cell $c$ and bin $b$ is calculated as:
\begin{equation}
    H(c,b) = \sum_{(x,y) \in c} [g(x,y) \in b],
\end{equation}

where $[g(x,y) \in b]$ is an indicator function that returns 1 if the gradient direction of the pixel $(x,y)$ falls within the range of bin $b$, and 0 otherwise.

Finally, the feature vector representing the image is formed by concatenating the histograms for all cells:
\begin{equation}
    f = [H(c_1), H(c_2), \dots, H(c_n)],
\end{equation}

where $n = |C|$ is the number of cells. The feature vector $f$ is the hog feature.

\end{document}